\documentclass[sigconf]{acmart}

\usepackage{algpseudocode}
\usepackage{algorithmicx}
\usepackage[ruled]{algorithm2e}
\usepackage{multirow}
\usepackage{subfigure}

\AtBeginDocument{%
  }

\setcopyright{acmcopyright}
\copyrightyear{2018}
\acmYear{2018}
\acmDOI{XXXXXXX.XXXXXXX}

\acmConference[Conference acronym 'XX]{Make sure to enter the correct
  conference title from your rights confirmation emai}{June 03--05,
  2018}{Woodstock, NY}

\acmPrice{15.00}
\acmISBN{978-1-4503-XXXX-X/18/06}

\begin{document}

\title{Origin-Destination Network Generation via Gravity-Guided GAN}



\author{Can Rong}
\affiliation{%
  \institution{Tsinghua University}
  \city{Beijing}
  \country{China}
}
\email{rc20@mails.tsinghua.edu.cn}

\author{Huandong Wang}
\affiliation{%
  \institution{Tsinghua University}
  \city{Beijing}
  \country{China}
}
\email{wanghuandong@tsinghua.edu.cn}

\author{Yong Li}
\affiliation{%
  \institution{Tsinghua University}
  \city{Beijing}
  \country{China}
}
\email{liyong07@tsinghua.edu.cn}

\begin{abstract}
  Origin-destination (OD) flow, which contains valuable population mobility information including direction and volume, is critical in many urban applications, such as urban planning, transportation management, etc. However, OD data is not always easy to access due to high costs or privacy concerns. Therefore, we must consider generating OD through mathematical models. Existing works utilize physics laws or machine learning (ML) models to build the association between urban structures and OD flows while these two kinds of methods suffer from the limitation of over-simplicity and poor generalization ability, respectively. In this paper, we propose to adopt physics-informed ML paradigm, which couple the physics scientific knowledge and data-driven ML methods, to construct a model named \textbf{O}rigin-\textbf{D}estination \textbf{G}eneration \textbf{N}etworks (\textbf{ODGN}) for better population mobility modeling by leveraging the complementary strengths of combining physics and ML methods. Specifically, we first build a \textbf{M}ulti-view \textbf{G}raph \textbf{A}ttention \textbf{N}etworks (\textbf{MGAT}) to capture the urban features of every region and then use a gravity-guided predictor to obtain OD flow between every two regions. Furthermore, we use a conditional GAN training strategy and design a sequence-based discriminator to consider the overall topological features of OD as a network. Extensive experiments on real-world datasets have been done to demonstrate the superiority of our proposed method compared with baselines.
\end{abstract}





\maketitle

\section{Introduction}

Human mobility is an essential component of the urban system and has a critical influence on the efficiency of city management \cite{simini2021deep}. 
However, real-world human mobility data is not always available for privacy issues and data collection costs (i.g., the cost of deployed sensors).
Thus, generating human mobility based on urban structures, named origin-destination (OD) generation in this literature, becoming a promising solution.
Specifically, OD flows, which contains the information of directed population flow between every two regions in the city, describes the human mobility in the crowd perspective, which plays a critical role in intelligent transportation systems~\cite{Sobral2021Knowledge, deng2016latent}.
It is also worth noting that, different from the traditional OD prediction or inference problem, the OD generation problem intends to generate OD information for a city of interest where no OD information can be obtained, which is a much more challenging task.

Although the study of population flow has been going on since a remarkably early time, there is still a lack of effective solutions for OD generation. Existing works can be summarized into two categories. One of them is knowledge-driven physical modelling, such as gravity models \cite{barbosa2018human,lenormand2016systematic}, radiation models \cite{simini2012universal} and intervening opportunities \cite{ruiter1967toward}, etc. This class of approaches usually analogizes complex human movement to simple physical laws. For example, the gravity model analogizes population flow to gravitational forces between celestial bodies \cite{barbosa2018human,lenormand2016systematic}, and the radiation model analogizes human mobility to radiation emission and absorption processes in solid-state physics \cite{simini2012universal,kittel2018kittel}.
The second category of approaches is data-driven models based on machine learning (ML) techniques. Recently developed powerful ML models, such as random forest and neural networks, can be used to fit arbitrary functions to characterize the dependency between variables. They have constructed complex models and achieved state-of-art on fitting the training distribution. Robinson et al.~\cite{robinson2018machine} leverage Gradient Boost Regression Trees (GBRT) and Pourebrahim et al.~\cite{pourebrahim2019trip} use the random forest to map the pair-wise regional attributes to the population flow between two regions. Simini et al. \cite{simini2021deep} extend the naive gravity model to deep residual neural network. Liu et al. \cite{liu2020learning} propose to learn a geo-contextual embedding by graph attention networks (GAT)~\cite{velivckovic2017graph} before utilizing GBRT to make a prediction. Nonetheless, both categories have their inherent limitations and therefore cannot model population mobility very well.

The physical models are incomplete by only adopting simple physical laws to model the relation between the population mobility and a limited number of factors, while human movement behavior is complex and affected by multiple factors including diverse region attributes, transport infrastructures, etc., which is ignored in the simple physical laws. This incompleteness leads to the disability of the physical models in terms of capturing the complex relations between the urban structure and human movement behavior, which intrinsically causes the under-performance of the physical models.
Meanwhile, the data-driven methods focus on using ML models to directly fit the training data without utilizing any prior knowledge. Thus, this category of methods is easy to incorrectly fit the unique mobility patterns or noise of the training dataset instead of capturing the intrinsic law of population movements, leading to their low generalization capability and bad performance on samples not included in the training data.
Furthermore, OD flows could be considered as a large-scale weighted network from the network view. But none of the works mentioned above take the topology of OD network into consideration and have a holistic grasp of population mobility in the city, which is also important of constructing OD networks \cite{saberi2017complex}.

In recent years, the rising paradigm of physics-informed ML~\cite{willard2020integrating} provides us a promising solution to build a powerful OD flow generation model by coupling the physics scientific knowledge and data-driven ML methods. 
By integrating the advantages of intrinsic and universal physics laws and ML models with strong fitting capabilities, this paradigm is able to overcome the disadvantages of incomplete physics laws and the low generalization capability of ML models.
Numerous existing literatures have reported successful results in various application area using physics-informed ML modelling, such as earth systems \cite{reichstein2019deep}, climate science \cite{faghmous2014big,krasnopolsky2006complex,o2018using}, turbulence modelling \cite{bode2019using,mohan2018deep,xiao2019reduced}, material discovery \cite{cang2018improving,raccuglia2016machine,schleder2019dft}, biological sciences \cite{yazdani2020systems} and etc.
Thus, we propose to construct a physics-informed ML model to leverage the complementary strengths of combining traditional physics knowledge and ML methods, and break the limitation of using only one of them when solving the OD generation problem. 
With the help of the physics-informed ML paradigm, ML models and physics laws are mutually complementary combined, exploiting knowledge to compensate for the limitations of poor generalization performance of ML models, and exploiting ML to compensate for the limitations of incomplete physical laws. Therefore, using physics-informed ML modeling to solve the OD generation problem is a promising direction.

Although there have been published works on both OD generation filed and physics-informed ML, it remains three challenges utilizing physics-informed ML to solve the OD generation problem. 
\textit{\textbf{First}}, there is a natural gap between physics law and data-driven ML models, which are quite different in form. The laws of physics often have a small number of interpretable parameters while ML models will have a huge number of uninterpretable parameters, so the two cannot be integrated directly.
\textit{\textbf{Second}}, the city is inherently a very complex system. There are diverse factors affecting human mobility in cities, both in terms of a wide range of regional attributes and various transportation networks. 
\textit{\textbf{Third}}, the network topological features are essential \cite{saberi2017complex} but abstract. It is difficult to find explicit metrics to evaluate the disparities between two directed weighted networks that can be used to train the model.

In this paper, we propose \textit{\textbf{ODGN}} (Origin-destination Generation Networks), a novel physics-informed ML model training with conditional GAN (generative adversarial networks) \cite{mirza2014conditional} framework, to combine the traditional physics knowledge and data-driven ML methods to solve OD generation problem well. The three challenges mentioned above are figured out through three special designs. \textit{\textbf{First}}, we propose a physics-informed encoder-decoder framework to couple the gravity law with data-driven methods, where the encoder is a neural network based spatial feature extraction structure and the decoder is a predictor inspired by the gravity law. The model is trained to learn the relationships between variables from a large amount of unstructured data, while the optimization direction of the model is correctly constrained by universal physics laws. \textit{\textbf{Second}}, we design the multi-view graph attention networks (MGAT) to capture diverse regional attributes and various transportation networks into our model to comprehensively model a city. \textit{\textbf{Finally}}, we introduce conditional GAN \cite{mirza2014conditional} to generate the OD network of a city given regional attributes and complete transportation topology, where the discriminator is used to distinguish the real OD network from the generated fake OD network, so that our model can capture the overall topological characteristics of the OD network.

Our contributions can be summarized as follows:
\begin{itemize}
\item  We are the first to propose utilizing physics-informed ML methods, which integrate the gravity law and data-driven neural networks, to solve the OD generation problem to the best of our knowledge.
\item  We design a novel physics-informed ML model called ODGN, which consists of MGAT and gravity-guided predictor to comprehensively model the population movement in the complex city system.
\item  We introduce the conditional GAN framework as our training strategy. We specially designed a random walk sampling-based network classification discriminator to consider the topological features of OD networks.
\item  Extensive experiments have been done to prove the superior performance of our proposed method compared with the state-of-art.
\end{itemize}

\section{Preliminaries} \label{sec:preliminaries}
In this section, we give a systematic introduction to the definition of necessary notations and give the problem formulation of OD generation.

\subsection{Definitions} \label{Sec:notation}

Before diving into the methods, some necessary definitions and notations should be presented first.

\textbf{Definition 1 (City)} In this literature, one city is used to indicate a large area, specifically a spatial space for human activity. We denote a city as $C \in \mathcal{C}$. And $\mathcal{C}^{train}$ is used to stand for the cities whose data is used for training the models while $C^{target}$ is used to represent the target city for which we want to generate OD information.

\textbf{Definition 2 (Region)} A region describes a small area of the territory included in the entire geographical space of a city. A city consists of $N$ regions, and there is no overlap between any two regions. We denote a region as $r \in \mathcal{R}$, where $\mathcal{R}$ means the set of regions, and use superscripts to indicate the city to which the region belongs, such as $r^{C^{target}}$.

\textbf{Definition 3 (Urban Regional Attributes)} Regions have attributes $X$ including demographic structure, economic indicators, POI (point of interest), etc. We use $x_i$ to denote the attributes of region $i$ as a feature vector.

\textbf{Definition 4 (Transportation Networks)} The spatial dependencies between regions within a city is shown as the way for people to get around via transportation, which is represented using $\mathcal{T}$. This work introduces regional neighborhoods, buses, and rail transit, three transportation-related urban spatial dependencies, i.g., $\mathcal{T}^{C^{train}}=\{T^{C^{train}}_{ngb}, T^{C^{train}}_{bus}, T^{C^{train}}_{rail}\}$.

\textbf{Definition 5 (Origin-destination Network)} Population mobility, consisting of thousands or millions of origin-destination trips, can be considered as a directed weighted network, $\mathcal{N}$, where the nodes stand for regions and the edges represent flows between pairs of regions.

\subsection{Problem Formulation}
The cities, $\mathcal{C}^{train}$, with OD network accessible will be used to train for OD network generation task in the target city $C^{target}$.

\textbf{Definition 6 (OD Network Generation)} Given the regional attributes $X$ as well as the Urban Graphs $\mathcal{G}$ of a city $\mathcal{C}$, we intend to generate the OD network $\mathcal{N}$ of the city $C$.

\section{Methodologies} \label{sec:methods}

\begin{figure*}[t]
  \centering
  \includegraphics[width=0.90\textwidth]{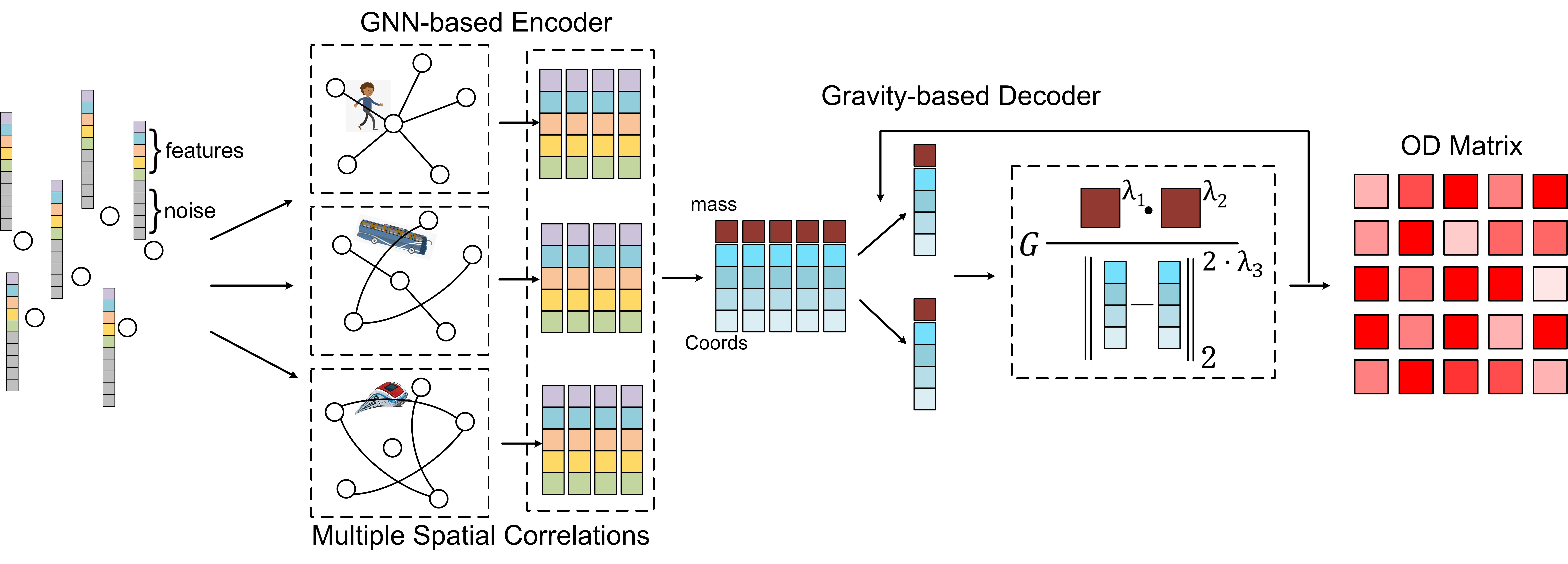}
  \caption{The architecture of OD networks generator.}
  \vspace{-0.2cm}
  \label{fig:generator}
\end{figure*}

In this section, we introduce the framework of our proposed method which is a persuasive and practical solution to the OD generation problem. The method includes a population mobility model called \textbf{ODGN} and a specially designed conditional GAN based training strategy. 

Figure \ref{fig:generator} shows an overview of ODGN, the physics-guided ML model, which consists of physics law modeling and data-driven neural networks. As we can see, the model contains two parts, a GNN based encoder and a gravity-based decoder. The encoder accepts regional attributes and transportation topologies as input, and output the node embedding, that captures the city's portrait. The decoder predicts OD flow between two regions given the embeddings of these two regions. Finally, the OD networks will be generated until OD flow between every two regions is obtained.

In our training strategy, there exist two phases. In the generation phase, the urban regional attributes and urban topology of the target city are fed into ODGN as conditions to generate fake OD networks. In the discrimination phase, we first apply a probability-based random walk on graph sampling strategy to sample sequences from the generated OD network and real OD networks, respectively. Afterward, a sequence modeling discriminator is utilized to distinguish whether sequences are sampled from real OD networks or generated ones.
The generator (ODGN) and the discriminator act as two rivals competing with each other: the generator tries to generate the fake OD networks under the supervision of the discriminator, while the discriminator aims to separate the real OD networks from fake ones to avoid got fooled by the generator. The adversarial learning phases drive both the generator and the discriminator to improve their performance, thus the generator could capture the ability to generate realistic OD networks given prior information of a city.

\subsection{Origin-destination Generation Networks}
In this part, we will give a detailed introduction to \textit{\textbf{ODGN}}, the physics-guided ML Modeling for population mobility. The whole structure is a encoder-decoder framework. 
The encoder is the multi-view graph attention networks, where different graphs are constructed by the different topology of transport modes. The graph construction will be detailed in the next section.
The decoder is a predictor of population flow between pair-wise regions of origin and destination. 

\subsubsection{Multi-view Graph Attention Networks.}
First, we should extract the features of the region include not only the regional attributes of its own, but also the features of regions that have strong spatial dependencies with it \cite{liu2020learning}. Existing works only take the distance into consideration and have neglected the extremely fast, and more important, public transportation system. We use multi-graphs to model the different transport modes to capture the comprehensive spatial dependencies between regions in the city. The nodes stand for regions and the edges mean spatial dependencies between regions.

As shown in Fig. \ref{fig:generator}, we constructed three graphs to model the spatial association of regional spatial neighborhoods, direct bus access and direct subway access in the city, respectively. The adjacency matrix of these three graphs are denoted as $A_{ngb}$, $A_{bus}$ and $A_{rail}$, where ${A_{rail}}_{i,j}=1$ means there lies a direct rail dependency between regions $r_i$ and $r_j$.

The graph convolutional layers we adopt are GAT (graph attention networks) \cite{velivckovic2017graph}, which employ attention mechanisms to aggregate the information from the neighbors of each node on graph. The computational formula of a GAT layer is shown as follow.
\begin{equation}
    {h'_i = \bigoplus_{k=1}^K} \sigma (\sum_{j\in {Neighbor}_i} \alpha^k_{ij}\textbf{W}^k h_j),
\end{equation}
where $h_j$ means the input node feature of region $r_j$ while $h'_i$ means the output feature map of region $r_i$, $K$ is the number of heads of \textit{multi-head attention}, $\alpha^k_{ij}$ means the $k^{th}$ head attention weight between region $r_i$ and region $r_j$, $\textbf{W}^k$ means the learnable weights, $\bigoplus$ denotes the concatenation and $\sigma$ is the activation function. The attention weights can be computed as the following formula according to \cite{velivckovic2017graph}.
\begin{equation}
{\alpha_{ij} = \frac{exp(LeakyReLU(\Theta[\textbf{W}h_i||\textbf{W}h_j]))}{\sum_{n\in\mathcal{N}}exp(LeakyReLU(\Theta[\textbf{W}h_i||\textbf{W}h_n]))}},
\end{equation}
where $\Theta$ and $\textbf{W}$ are all learnable weights, $||$ denotes concatenation.

We construct three kinds of graphs as mentioned above, each of which can obtain node embedding by several graph convolutional layers to grasp the corresponding transportation topology.
Then, we concatenate the outputs of the three graphs and apply a linear mapping to get the overall representation of regions, the process is shown below.
\begin{equation}
    E_i = (Emb^{ngb}_i||Emb^{bus}_i||Emb^{rail}_i)W,
\end{equation}
where $Emb_{bus}$ means the node embedding accessed by the graph neural networks with convolution of spatial dependencies of bus transportation, $W$ means the learnable weights and $E$ denotes the final representation of region $r_i$.
It is worth noting that node embedding can be learned by parameter optimization, subject to the constraints of the later decoder, and we make node embedding optimized in the designed direction.

\subsubsection{Gravity-based flow predictor.}
In order to improve the generalization of the proposed ML model and to correctly constrain the optimization direction of the model parameters, we propose to use the physical law of gravitation to strengthen our model. The gravity law of physics is proposed by Newton \cite{newton1833philosophiae}. According to the law of gravitation,  there is an attractive force between any two objects that is proportional to the mass of the two objects and inversely proportional to the square of the distance between them, as shown in the following equation.
\begin{equation}
    F = G\frac{m_im_j}{r^2}.
\end{equation}
In 1946, Zipf proposed that the population flow could be calculated by an equation motivated by the gravity law. In his work, Zipf points out that the population migration between two regions is proportional to the number of people of two regions and inversely proportional to the distance between them. The equation is shown below.
\begin{equation}
    T_{ij}  \propto \frac{P_iP_j}{r_{ij}},
\end{equation}
where $T_{ij}$ means the magnitude of migratory people flow, $P_{i}$ denotes the population number of region $r_i$ and $r_{ij}$ means the distance between regions $r_i$ and $r_j$. This work is the gravity model that is widely used in numerous applications. Recently, some works gives the gravity model several learnable parameters to accommodate the differences in the relationships of population distribution and population mobility under different urban structures. The final computation formula is as follows.
\begin{equation} \label{eq:gm}
    T_{ij} = G \frac{P_i^{\lambda_1}P_j^{\lambda_2}}{r_{ij}^{\lambda3}},
\end{equation}
where $G$ and $\{\lambda_{i} | i = 1,2,3\}$ are the learnable parameters to a specific city.

As can be seen, the law of gravity is very universal, therefore we adopt it to design the model and constrain the model parameters to optimize in the direction more consistent with the population movement theory. Inspired by Gravity-inspired Graph AE \cite{salha2019gravity}, we split the final node embedding capturing the regional attributes and multiple transportation topologies into two parts. One part characterizes the \textit{mass} of the region and the other part characterizes the \textit{location} of the region in the abstract feature space. The mass does not represent a physical mass, but rather a representation of the region's contribution to population flow production or attraction. Given the \textit{locations} of the regions, we could obtain the distance between every two regions by computing the \textit{Euclidean distance} based on the coordinates of the \textit{location} as shown in Fig. \ref{fig:generator}. Next, every element in the OD matrix could be computed by Eq. \ref{eq:gm}. We also accept the four learnable parameters $G$ and $\{\lambda_{i} | i = 1,2,3\}$ in Eq. \ref{eq:gm} to enhance the modeling ability of our model. Finally, the generated OD networks could be constructed as a directed weighted network based on the OD matrix computed by the multi-view graph attention networks and gravity-based flow predictor. The OD matrix is the adjacency matrix of the generated OD network.

\subsection{Conditional GAN-based Training Strategy}
\begin{figure}[t]
    \centering
    \includegraphics[width=0.45\textwidth]{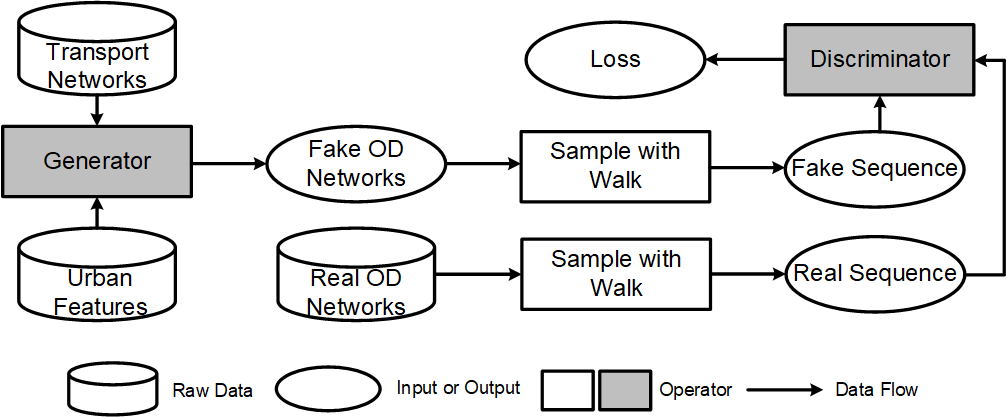}
    \caption{An overview of our Conditional GAN-based training strategy.}
    \label{fig:training}
\end{figure}

With the model structure in mind, we next introduce the specially designed Conditional GAN-based model training strategy. As shown in Fig. \ref{fig:training}, the training process includes two phases, generation and discrimination. In the generation phase, the generator generates the fake OD network given the regional attributes of every region and the transportation networks. The generator is our proposed ODGN model. In the discrimination phase, due to the limitation of computational and storage resources, we design a discrimination method based on random walk sampling sequences to distinguish the generated OD network and real OD network. For ease of presentation, we call generator $\mathcal{G}$ and discriminator $\mathcal{D}$.

We choose the Wasserstein GAN \cite{arjovsky2017wasserstein} as our framework. According to the original paper, the min-max game is optimized as the objective below.
\begin{equation} \label{eq:wgan}
    \max_{w\in \mathcal{W}} \mathbb{E}_{x\sim \mathbb{P}_r} [f_w(x)] - \mathbb{E}_{z\sim p(z)} [f_w(g_{\theta}(z))],
\end{equation}
where $\mathcal{W}$ denotes the weight space, $f$ means the discriminator networks and $g$ means the generator networks, i.g. ODGN, and $\theta$ means the learnable weights of generator networks. The optimization objective aims to reduce the Wasserstein distance between the generated data distribution and the real data distribution. The detailed introduction of the training strategy will be given in the following sections.

\subsubsection{\textbf{Generator}}
To generated OD networks with local pairwise regional features and global network topological features, the generator $\mathcal{G}$ we adopt is ODGN, which has been described above. The input of the $\mathcal{G}$ contains two parts, i) a random noise tensor $\mathcal{Z} = {z_1, ..., z_{|\mathcal{R}|}}$ (where $|\cdot|$ means the cardinality of a set) sampled from Gaussian distribution, . ii) a condition tensor $\mathcal{C} = {C_{r_1}, ..., C_{r_{|\mathcal{R}|}}}$ composed of regional attributes of all regions. The condition tensor $\mathcal{C}$ is concatenated into the noise tensor $\mathcal{Z}$ so that $\mathcal{G}$ construct the mapping from distribution $p_{\mathcal{Z}}(\mathcal{Z})$ to the OD network distribution $\mathcal{G}(\mathcal{C}, \mathcal{Z})$. It is also noteworthy that the multi-mode transportation networks, we denote $\mathcal{T}$ in the following, are used to construct the graph of MGAT. Although they are not the input to the neural network, they act as a condition for generating OD networks as well. Hence, the generator builds a mapping to $\mathcal{G}(\mathcal{C}, \mathcal{T}, \mathcal{Z})$ in the end.

\subsubsection{\textbf{Probability-based random walk sampling strategy}}
It is not easy to distinguish direct weighted networks, i.g. OD networks, directly. Inspired by NetGAN \cite{bojchevski2018netgan}, we design a method to sample from OD networks to get sequences, and distinguish the sequences from the generated OD network or the real ones. Different from NetGAN, the sampled sequences need to ensure both that the topology of the graph is captured and that the edge weight information, i.g. OD flow volumes, are preserved. Therefore, the probability that random walk travels between nodes of networks when sampling should be associated with the weights of edges. In this method, the probability of each edge being walked next is equal to the ratio of the OD flow volume represented by that edge to the current node out-degree, i.g. outflow of the region represented by the current node. The formula for calculating the probability can be written as follows.
\begin{equation} \label{eq:rw}
    P(Node_{r_i}|Node_{r_i}) = \frac{T_{ij}}{\sum_{k \in {Neighbor}_i} {T_{ik}}},
\end{equation}
where $P(Node_{r_i}|Node_{r_i})$ means the probability of walk from nodes stand for region $r_i$ to node stands for region $r_j$. The initial node of each sequence is selected with equal probability at random among all nodes of the OD network, and then a sequence of nodes of length $L$ is sampled by randomly walking $L$ times on the network according to Eq. \ref{eq:rw} and the initial node could be dropped after the sample of a sequence. We denote the sample procedure as the following formula.
\begin{equation}
    \mathcal{S} = ProbWalk(\mathcal{N}).
\end{equation}
To ensure that the sampling process does not interrupt the gradient backpropagation, similar to NetGAN \cite{bojchevski2018netgan}, we used the Straight-Through Gumbel estimator \cite{jang2016categorical} by Jang.

\subsubsection{\textbf{Discriminator}}
We should determine whether the sequences are sampled from the generated OD network or real OD networks. First, we need to decide what information is contained in the sequences that will provide adequate guidance for the discriminator $\mathcal{D}$. We choose the regional attributes $\mathcal{X}={x_1, ..., x_{L}}$ of the regions denoted by nodes in the sequence and OD flow volumes $\mathcal{T} = {t_1, ..., t_{L}}$ of edges walked. The regional attributes and OD flow volumes are concatenated as a final sequence $\mathcal{S}={(x_1||t_1), ..., (x_L || t_L)}$ of length $L$.

Then We choose TCN (Temporal Convolutional Networks) \cite{bai2018empirical}, which has been shown to be superior in modeling long sequences, as our sequence discriminator $\mathcal{D}$. More specifically, the TCN does layer-by-layer 1D-convolution on the sequence and expands the perceptive field of the convolutional kernels layer by layer by dilation convolution to obtain a global representation of the long sequence next. To meet the training requirements of the Wasserstein GAN, the discriminator $\mathcal{D}$ finally outputs a value through the TCN and an immediately connected fully connected layer. The computational procedure of the discriminator could be formulated as follows.
\begin{equation}
    \mathcal{D}(\mathcal{S}) = FC(TCN(\mathcal{S})),
\end{equation}
where $FC$ means the fully-connected layer.
As described in Eq. \ref{eq:wgan}, the sequences from generated OD network and real OD networks should be as different as possible from the output obtained by discriminator.

\subsubsection{\textbf{Training Algorithm}}
Then, we outline the training algorithm in detail as shown in Algorithm \ref{alg:Algorithm 1}. During the continuous iterative optimization, the generator and discriminator are simultaneously improved to obtain excellent generators. The iteration is stopped when the loss converges.

\begin{algorithm}[t]
\caption{Conditional GAN Training Algorithm}
\label{alg:Algorithm 1}
\LinesNumbered
    \KwIn{ \\
    \quad Regional features of the training city $\mathcal{X}^{C^{train}}$.\\
    \quad Transportation networks of the training city $\mathcal{T}^{C^{train}}$.\\
    \quad Regional features of the target city $\mathcal{X}^{C^{target}}$.\\
    \quad Transportation networks of the target city $\mathcal{T}^{C^{target}}$.\\
    \quad OD networks of the training cities $\{\mathcal{N}^{C^{train}}\}$.\\}
    \KwOut{\\
    \quad Learned ODGN model.
    }
    
    Construct the graphs according to transportation networks of the training and target city \;
    Initialize the learnable parameters of model \;

    \Repeat{Generation loss $loss_\mathcal{G}$ and $loss_\mathcal{D}$ converge}{
        \For{$i=1$ \textbf{to} $n_{critic}$}{
            Select one city from all cities and one real OD network $N_{real}$ from $\{\mathcal{N}^{C^{train}}\}$ \;
            Generate OD network $\mathcal{N}_{fake}$ of the selected city given its $\mathcal{X}$ and $\mathcal{T}$ by $\mathcal{G}$ \;
            Sample a batch of sequences $\mathcal{S}_{fake}$ from $\mathcal{N}_{fake}$ and sample a batch of sequences $\mathcal{S}_{real}$ from $\mathcal{N}_{real}$ \;
            Compute $loss_\mathcal{D} = -\mathcal{D}(S_{real}) + \mathcal{D}(S_{fake})$ \;
            $w_{\mathcal{D}} \leftarrow clip(w_{\mathcal{D}})$ \;
            Optimize $\mathcal{D}$ by $loss_D$ \;
        }

        Select one city from all cities \;
        Generate OD network $\mathcal{N}_{fake}$ of the select city given its $\mathcal{X}$ and $\mathcal{T}$ by $\mathcal{G}$ \;
        Compute $loss_\mathcal{G} = - \mathcal{D}(S_{fake})$
        Optimize $\mathcal{G}$ by $loss_G$
    }
\end{algorithm}

\section{Experiments} \label{sec:exp}
In this section, we conduct  systematic experiments on real-world datasets of American cities  to answer the following questions:

\begin{itemize}
 \item \textbf{RQ1:} Can our proposed physics-informed ML model ODGN with a specially designed training strategy effectively generate OD networks?
 \item \textbf{RQ2:} Does the design of each part of the model work?
 \item \textbf{RQ3:} What does the mass of the region learned by the model indicate?
\end{itemize}
We will first introduce the experiment settings below, and then answer the above research questions.

\subsection{Experiment Settings}
\subsubsection{Datasets and Preprocessing}
 We choose 8 cities, i.g. New York City, Los Angeles, Chicago, Houston, San Francisco, Seattle, Washington D.C., Memphis, in the United States for the experiment, and selected a large developed city (New York City), a second-tier city (Seattle) and a developing city (Memphis) for the OD generation experiment respectively, and the results proved the superiority of our proposed methodology. The data we use is completely public, and the data used will be introduced in several parts as follows. 
 \begin{itemize}
 \item \textbf{Demographics.} Demographic features are an important part of regional attributes. We can obtain information on regional demographic attributes down to the spatial granularity of the census block from the publicly available ACS (American Community Survey) project \cite{census2020} on the U. S. Census Bureau website. And we filtered 24 dimensions of these features as the demographic attributes of the region, including the number of people by gender and age, as well as by education and economic level.
 \item \textbf{POIs.} We crawl POIs data for each city from OpenStreetMap \cite{OpenStreetMap}, a crowdsourced, publicly available geographic information data site.
 \item \textbf{Transportation.} We retrieved data on each city's public transportation routes from their government websites or proprietary data sites.
 \item \textbf{OD flows.} The OD data we use is commuting data from the Longitudinal Employer-Household Dynamics Origin-Destination Employment Statistics (LODES) project.
 \end{itemize}
 
\textbf{Data Preprocessing.} We use the spatial granularity of census tracts for our study, with each census tract in a city being a region. We sum the demographic regional attributes of all the census blocks in that census tract. All POIs are allotted to the region in which it is located, and then the POIs in each region are counted according to 36 categories to get the POI distribution for each region. The demographic attributes of each region are concatenated with the POI distribution to form regional attributes. Transportation networks are used to construct the multi-graphs in MGAT. In the neighbor graph, an edge is constructed if two regions are adjacent to each other; in the bus related graph, an edge is constructed if two regions are directly connected by a bus route; in the railway related graph, an edge is constructed if two regions are directly connected by the subway. OD data are aggregated between every pair of census tracts.
 
\subsubsection{Metrics}
We adopt $\textit{F-JSD}$ (Flow Jensen-Shannon Divergence), ${\textit{RMSE}}$ (Root Mean Square Error) and $\textit{CPC}$ (Common part of Commute) as our evaluation metrics. $\textit{F-JSD}$ is the Jensen-Shannon Divergence between real OD flow data and generated OD flow data and is an evaluation of the overall distribution of the generated data. ${\textit{RMSE}}$ is an element-wise error evaluation of the flow volume on the generated OD network. $\textit{CPC}$ is the commonly used evaluation metric \cite{pourebrahim2019trip,robinson2018machine,liu2020learning,simini2021deep} for OD generation tasks.
 
 \begin{equation}
     \textit{F-JSD}{\;}={\;} \sqrt{\frac{\textbf{KL}(\textbf{P}_{\mathcal{N}_{fake}}||\textbf{P}_{\mathcal{M}}) +  \textbf{KL}(\textbf{P}_{\mathcal{M}}||\textbf{P}_{\mathcal{N}_{real}})}{2}},
 \end{equation}
 \begin{equation}
     {\textit{RMSE}}{\;}={\;}{\sqrt{{\frac{1}{|T|}}{\sum_{i,j}}{||}{{{T}_{ij}}-{\hat{T}_{ij}}}{{||}_2^2}}},
 \end{equation}
 \begin{equation}
     {\textit{CPC}}{\;}={\;} {\frac{2 \sum_{i,j} min(T_{ij}, \hat{T_{ij}}) } {\sum_{i,j}{\hat{T_ij}}+ \sum_{i,j} {T_{ij}} }},
 \end{equation}
 where $\textbf{KL}$ means the KL divergence, $\textbf{P}_{\mathcal{N}_{real}}$ means the distribution of the real data, $\textbf{P}_{\mathcal{N}_{fake}}$ means the distribution of the generated data, $\textbf{P}_{\mathcal{M}}$ means the element-wise mean of the $\textbf{P}_{\mathcal{N}_{real}}$ and $\textbf{P}_{\mathcal{N}_{fake}}$.
 
 \subsubsection{Baselines}
 \begin{itemize}
 \item \textbf{Gravity Model.} Gravity model \cite{lenormand2016systematic} is widely used in many applications and is inspired by Newton's law of Gravitation. Zipf correlates the gravity model with population mobility and made an outstanding contribution \cite{zipf1946p}.
 \item \textbf{Random Forest.} Random forest is a strong generalized data-driven model. Pourebrahim et al. \cite{pourebrahim2019trip} reported that it works as the state-of-art on population mobility modeling task.
 \item \textbf{Gradient Boosting Regression Tree.} GBRT \cite{robinson2018machine} combines the ensemble learning and gradient boosting technique to enhance the generalization capability of regression trees. 
 \item \textbf{Deep Gravity.} In order to enhance the complexity of the gravity model to allow modeling of more factors, Simini et al. utilize deep neural networks to extend the gravity model to deep gravity \cite{simini2021deep}.
 \item \textbf{GAT.} This baseline uses graph attention networks \cite{velivckovic2017graph} to extract spatial associations between regions to improve prediction accuracy.
 \item \textbf{GMEL.} Liu et al. propose GMEL \cite{liu2020learning}, which learn the geo-contextual embeddings first and predict the OD flow by GBRT.
 \item \textbf{GAT-GAN.} This baseline is to remove all the designs of our model and simply use GAT and GAN to generate OD networks.
 \end{itemize}

\subsubsection{Parameter Settings}
 In this section, we give a complete interpretation of the parameter settings of our model and all baselines.
 The $n_estimators$ of tree-based models, i.g. random forest and GBRT, is set to 100. The same with the original paper of deep gravity \cite{simini2021deep}, the number of hidden layers is set as 15. The number of graph convolutional layers is set to 3, the number of channels is set to 64, and the number of heads is set to 8 for all graph neural networks related models. In our model, our embedding size is chosen to be 64, $n\_critic$ is specified to be 5 for the first 300 epochs and 1 thereafter, and the noise dimension is set to be 60, the same as the dimension of regional attributes.

\subsection{Overall Performance (RQ1)}
 
 \begin{table*}[]
 \begin{tabular}{l|rrr|rrr|rrr}
 \hline
 \multirow{2}{*}{} & \multicolumn{3}{c|}{New York City}                                              & \multicolumn{3}{c|}{Seattle}                                                    & \multicolumn{3}{c}{Memphis}                                                    \\ \cline{2-10} 
                   & \multicolumn{1}{c}{F-JSD} & \multicolumn{1}{c}{RMSE} & \multicolumn{1}{c|}{CPC} & \multicolumn{1}{c}{F-JSD} & \multicolumn{1}{c}{RMSE} & \multicolumn{1}{c|}{CPC} & \multicolumn{1}{c}{F-JSD} & \multicolumn{1}{c}{RMSE} & \multicolumn{1}{c}{CPC} \\ \hline
 Gravity Model     & 0.4979                    & 12.74                    & 0.3547                   & 0.5193                    & 24.37                    & 0.3485                   & 0.5582                    & 27.05                    & 0.3155                  \\
 GBRT              & 0.2569                    & 7.70                     & 0.5600                   & 0.2846                    & 17.41                    & 0.6461                   & 0.2961                    & 19.87                    & 0.5854                  \\
 Random Forest     & 0.2520                    & 8.43                     & 0.5821                   & 0.2594                    & 16.59                    & 0.6896                   & 0.3022                    & 14.52                    & 0.6024                  \\ \hline
 Deep Gravity      & 0.2755                    & 8.86                     & 0.5633                   & 0.3099                    & 16.58                    & 0.6189                   & 0.2900                    & 16.80                    & 0.5640                  \\
 GAT               & 0.2540                    & 7.03                     & 0.5726                   & 0.2642                    & 21.63                    & 0.6510                   & 0.3011                    & 16.26                    & 0.5466                  \\
 GMEL              & 0.2540                    & 6.49                     & 0.5810                   & 0.2717                    & 21.42                    & 0.6510                   & 0.3013                    & 15.97                    & 0.5729                  \\
 GAT-GAN           & 0.2512                    & 5.99                     & 0.6017                   & 0.2508                    & 16.20                    & 0.6996                   & 0.2556                    & 16.42                    & 0.6147                  \\ \hline
 ours              & \textbf{0.2312}           & \textbf{5.20}            & \textbf{0.6476}          & \textbf{0.2400}           & \textbf{14.99}           & \textbf{0.7300}          & \textbf{0.2348}           & \textbf{14.00}           & \textbf{0.6282}         \\ \hline
 \end{tabular}
 \caption{Overall performance on real world datasets. The lower the F-JSD, RMSE and the higher the CPC, the better the performance.}
 \label{tab:performace}
 \end{table*}
 
We will present the performance comparison of our proposed method with all baselines in this section. As can be seen in Table \ref{tab:performace}, our proposed method achieves the best performance on all metrics, further demonstrating the advantage of the paradigm of physics-informed ML for the problem of OD generation. Table 1 shows the performance of our method and baselines on the 3 cities. From the experimental results in New York City, we can see that the traditional knowledge-based physics law performs poorly because it is too simple to capture complex human mobility patterns. Data-driven approaches including tree-based models and neural networks achieve better performance relative to the gravity model, but do not achieve the best performance due to the limitations mentioned in the previous section. Tree models, i.g. random forest and GBRT, have a more stable performance compared to graph neural network models, i.g. GAT, GMEL, GAT-GAN, thanks to the stability brought by ensemble learning, which improves the generalization ability. GMEL \cite{liu2020learning} adds the training strategy of multi-task learning, which gives certain constraints on the optimization direction from the loss and can improve the training effect of the model by a small margin. GAT-GAN shows suboptimal performance, indicating that it makes sense to consider the topological features of the OD network for the city. Our proposed method uses the paradigm of physics-informed ML, which combines the advantages of the universality of physics laws and the strong modeling ability of data-driven method, and uses the training strategy of GAN to capture the topological features of OD networks using a specially designed discriminator, so that the optimal performance is achieved. Performance on Seattle and Memphis is consistent with New York City, but poorer overall on RMSE. It is possible that this is due to the different urban structures between the cities, as Seattle and Memphis are relatively different from the large-scale cities used for training. However, our proposed method still achieves the best performance on all metrics among all methods and remains stable.

\subsection{Ablation Study (RQ2)}
 
 \begin{figure}[htbp]
 \centering
 \subfigure[Ablation study on $CPC$.]{
     \label{ablationcpc}
     \includegraphics[width=0.227\textwidth]{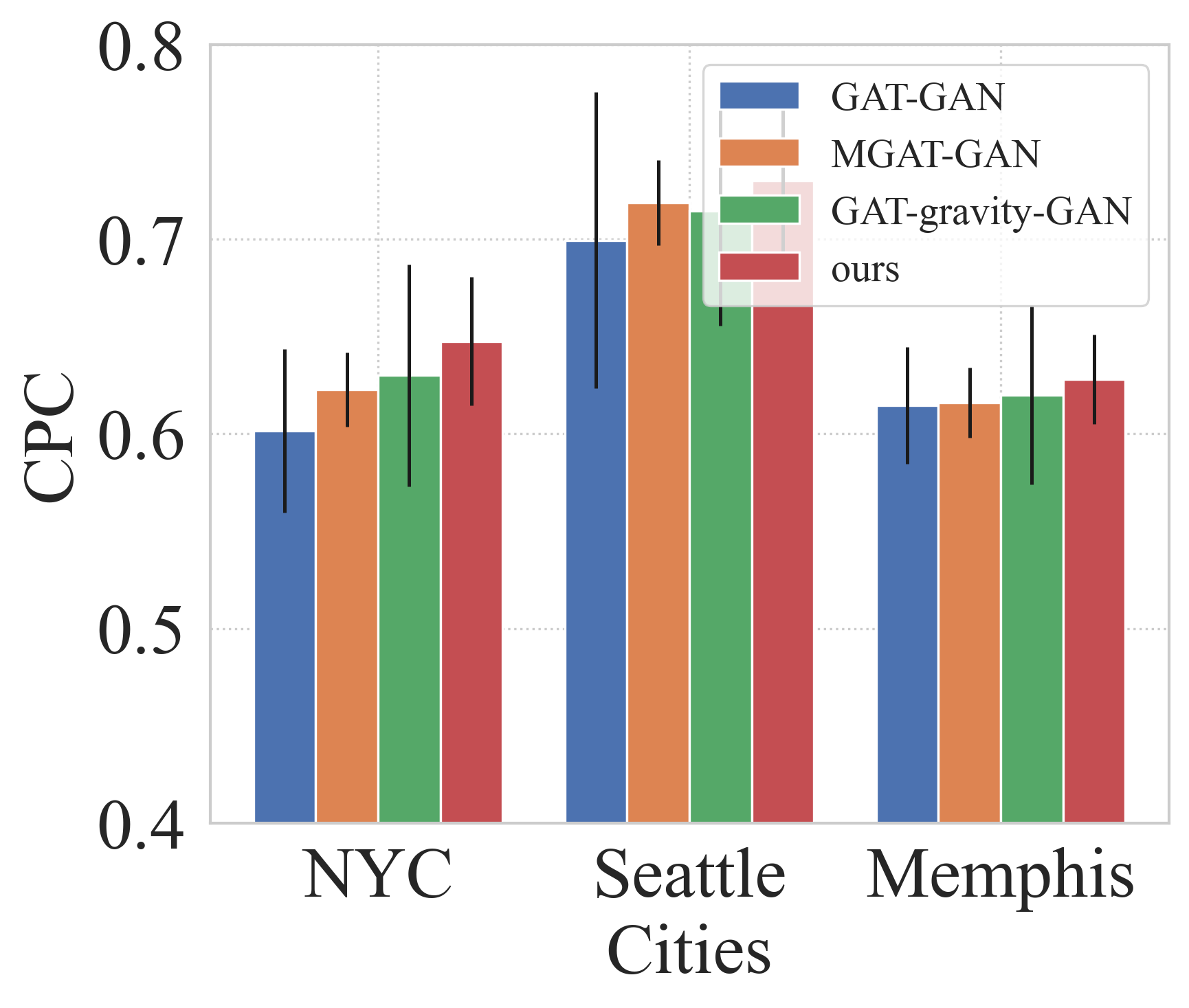}
 }
 \subfigure[Ablation study on $RMSE$]{
     \label{ablationrmse}
     \includegraphics[width=0.227\textwidth]{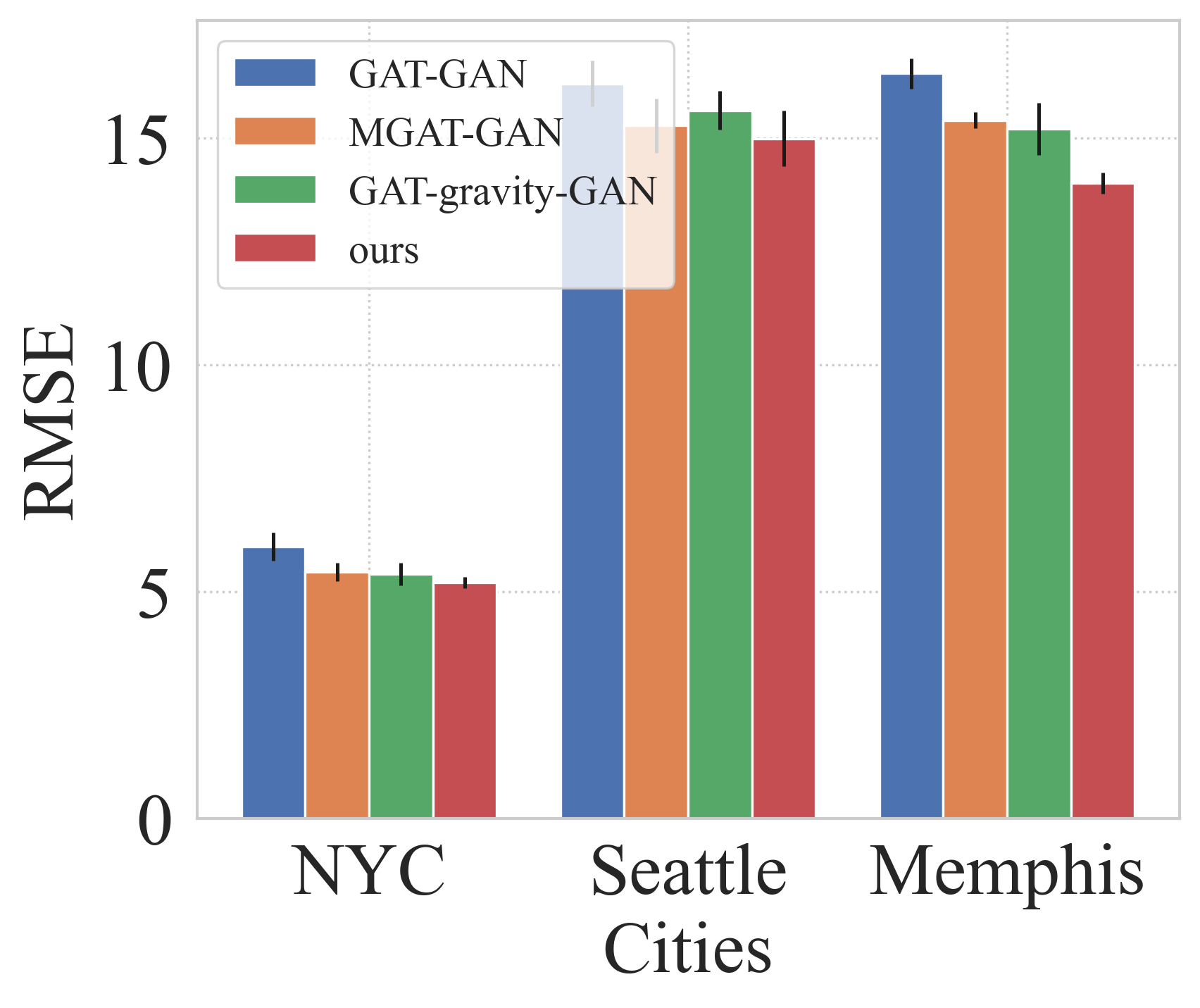}
 }
 \caption{Ablation study.}
 \label{fig:ablation}
 \end{figure}

In this section, we will carefully analyze the results of the ablation study shown in Fig. \ref{fig:ablation} and point out the reasons for the superior performance of our method. The basic model is GAT, but GAT is not repeated in this section since it differs from the other design parts of this method in terms of predictive and generative, and has been shown in Table \ref{tab:performace} to use conditional GAN in a way that considers the advantages of OD network topological features. This section focuses on the design of structure inspired by the gravitation law and the role of multi-view graph in the framework of conditional GAN training, which is detailed following one by one.
 
\textbf{Multi-view Graph Attention Networks}.
We first evaluate the effect of introducing comprehensive transportation networks by MGAT. As shown in Fig. \ref{fig:ablation}, from the metric of $CPC$ and $RMSE$, the design of adding the multi-view graph can improve the quality of generated OD data in all three cities. Only the incorporation of multi-graph design to introduce information about multiple transportation networks can bring about a 3\% performance improvement. After MGAT, we can get more informative node embedding to improve the quality of the final generated OD.
 
\textbf{Gravity-based Decoder}. 
After obtaining embeddings, the recent works often use bilinear or dot approach to predict the population flow between regions, which requires the relevant weight matrix $W$ to be able to learn the mapping from regional features to flows directly without guidance. However, this is very difficult, so we designed the decoder inspired by the law of gravity to reduce the number of parameters while using prior knowledge to constrain the optimization direction and improve the performance of the decoder. From the results shown in Fig. \ref{fig:ablation}, the design of merely gravity-based decoder can bring about a 4\% performance improvement.
 
\textbf{Ours}
We can bring more than 7\% performance improvement by adding both parts of the design to the model, which shows that the two parts of the design are not completely redundant. It is also worth noting that our proposed conditional GAN-based training strategy with discriminator design, as seen by comparing with GAT, can also bring a 3\% performance improvement alone. The combination of all the designs can bring a 13\% performance improvement.

\subsection{Explainability Study (RQ3)}
 \begin{figure}[htbp]
 \centering
 \subfigure[Scatter plot of population vs. mass.]{
     \label{masspop}
     \includegraphics[width=0.227\textwidth]{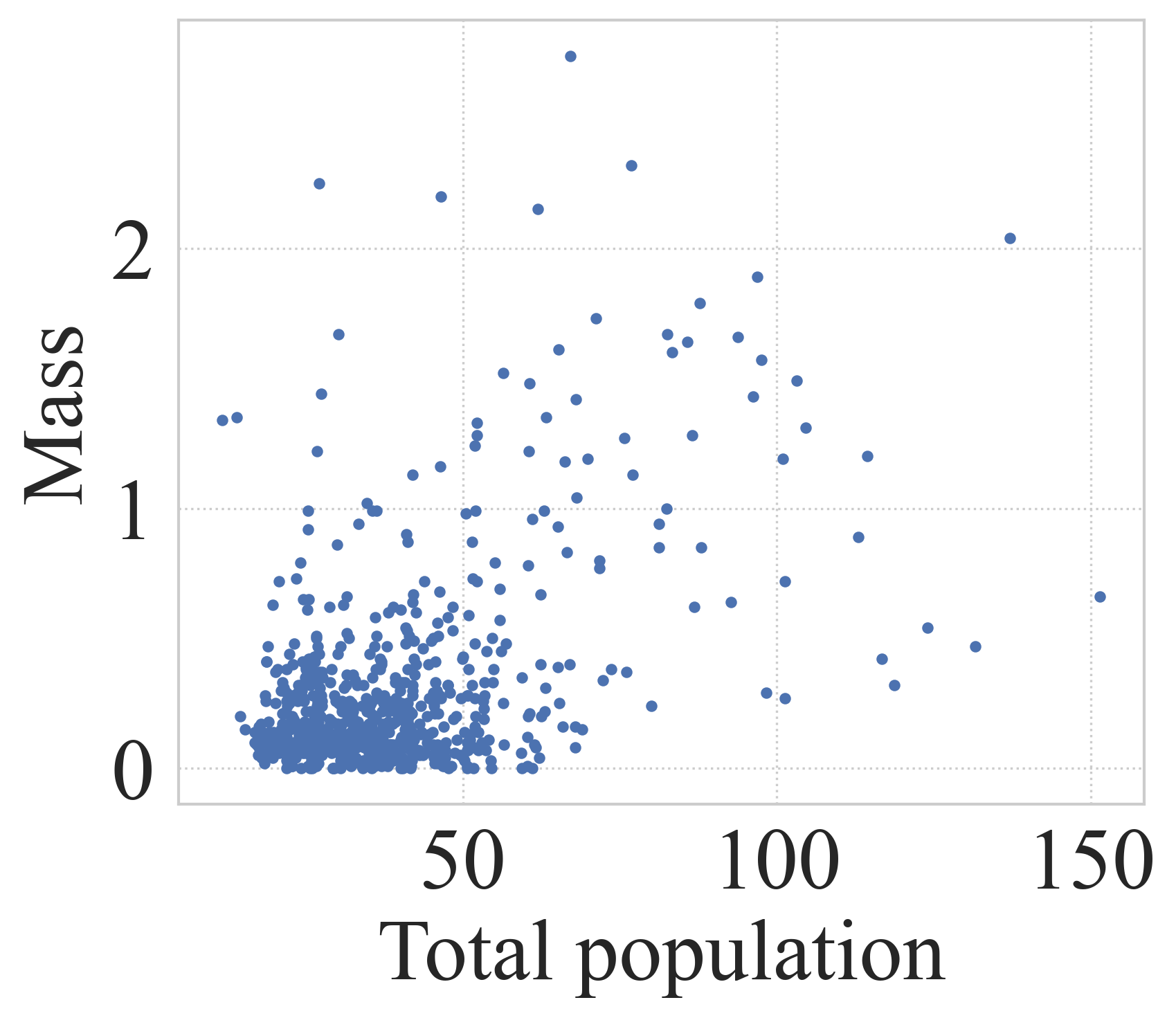}
 }
 \subfigure[Scatter plot of number of POIs vs. mass.]{
     \label{masspoi}
     \includegraphics[width=0.227\textwidth]{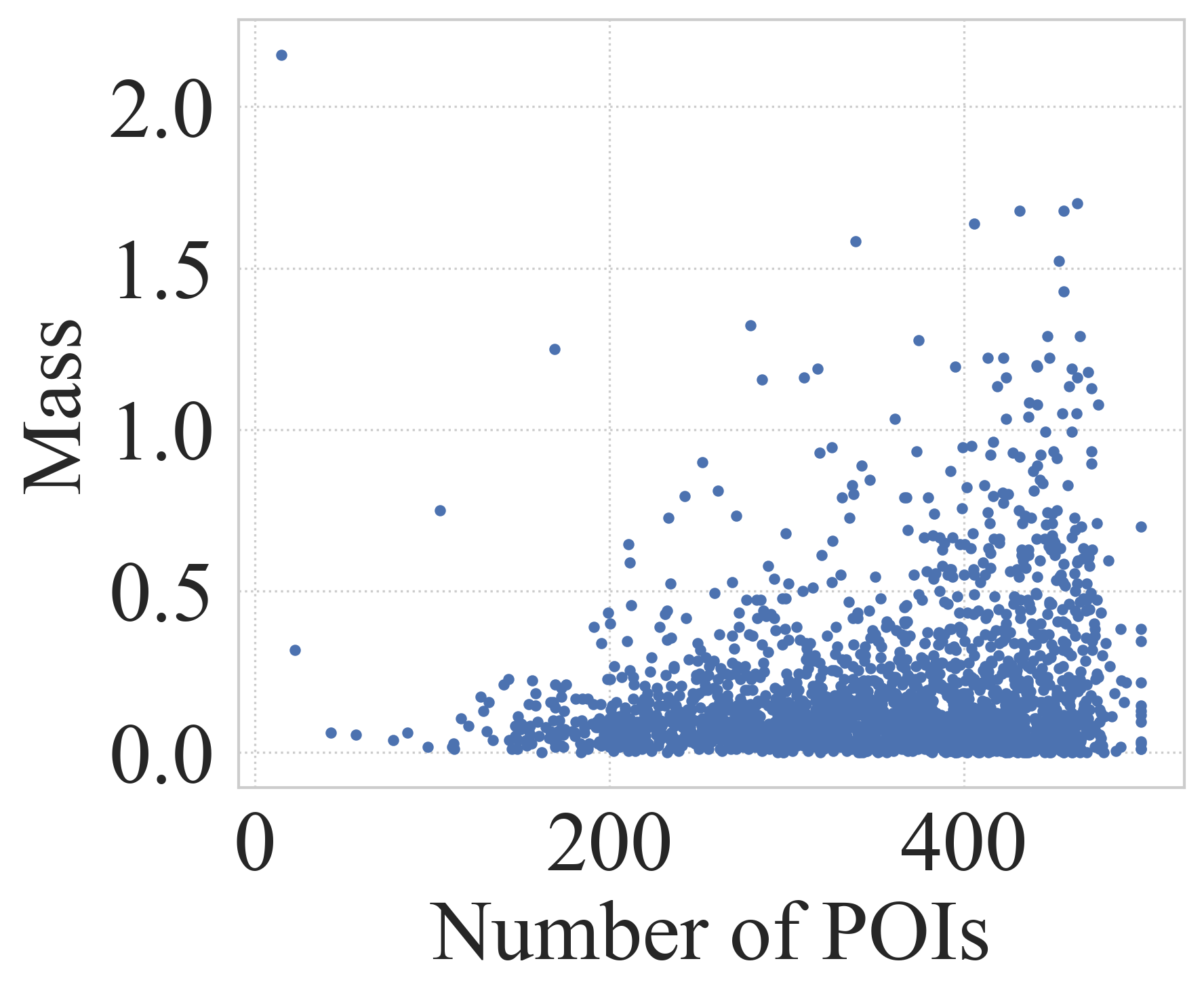}
 }
 \vspace{-10pt}
 \caption{Study on the explainability of gravity-based decoder.}
 
 \label{explainable}
 \end{figure}
 
In this part, we explore whether node embedding can actually learn meaningful information based on the guidance of the gravity-based design in the decoder, focusing on the mass of the representational region. To get a complete picture of what mass learned, we examined the correlation between the population number and the total number of POIs, which characterize the economic volume with respect to the value of regional mass, respectively. As can be seen in Fig. \ref{explainable}, mass learned to characterize the demographics as well as the representation of the economy. Specifically, the correlation between mass and population number and POI number is 0.47 and 0.20, respectively, after excluding some obvious noise region (population number is 0 but POI number is not 0). And it can be seen from Fig. \ref{masspoi} that there is a certain nonlinear constraint relationship between the POI and the learnable value of the mass.

\section{Related Work} \label{sec:Relatedwork}

\subsection{OD Flow Models}
Due to the vital role of OD flow in intelligent transportation systems in terms of numerous applications including traffic dispatching, transportation planning and travel routing \cite{castiglione2015activity}, the OD flow model has been widely studied, which can be divided into two categories, i.e., the knowledge-driven models and the data-driven models.
Representative knowledge-driven models include the gravity models \cite{barbosa2018human,lenormand2016systematic}, radiation models \cite{simini2012universal} and intervening opportunities \cite{ruiter1967toward}, etc.
These models are built by analogizing the population flow as classical physical mechanisms or processes.
For example, the gravity models are derived through analogy with the gravity law in Newtonian mechanics~\cite{barbosa2018human,lenormand2016systematic},
and the radiation models are derived through analogy with the radiation emission and absorption processes in solid-state physics~\cite{kittel2018kittel}.
In these models, the physics formulas describing gravity or radiation are utilized as the knowledge to describe the population flow.
Since the potential common principles between human mobility and different physical mechanisms or processes, these models are usually robust and maintain a stable performance on different cities. 
However,  they are incomplete with numerous ignored urban structure features including diverse regional attributes and various transportation networks, which have a large impact on human mobility behavior and thus cause the under-performance of these knowledge-driven models.

Meanwhile, the data-driven models leverage the strong modeling ability of machine learning methods to directly learn the distribution of OD flow from data correlated with urban structure features. Although most existing approaches focus on predicting future OD flow~\cite{spadon2019reconstructing, wang2019origin, shi2020predicting}, many OD flow generation methods have been proposed~\cite{robinson2018machine,pourebrahim2019trip,liu2020learning,simini2021deep,Yao2020Spatial}.
Specifically, Robinson et al. \cite{robinson2018machine} and Pourebrahim et al.~\cite{pourebrahim2019trip} leverage GBRT and random forest model to generate OD flow based on regional attributes, respectively. Liu et al.~\cite{liu2020learning} propose to learn the embedding features based on geo-adjacency network by GAT~\cite{velivckovic2017graph} combined with GBRT to predict future OD flow. Simini et al. \cite{simini2021deep} exploit many urban structure features (e.g., land use, road network, transport) by performing a concatenation of them and then utilize a feed-forward neural network to predict the OD flow.
Yao et al.~\cite{Yao2020Spatial} consider regional attributes including position, propulsiveness, and attractiveness, and then utilize an encoder-decoder neural network to generate OD flow.
This category of method is easy to incorrectly fit the unique mobility patterns or noise of the training dataset instead of capturing the intrinsic law of population movements, leading to their low generalization capability and bad performance on samples not included in the training data.
Different from these existing methods, our work focus on building a powerful OD flow generation model by coupling the physics scientific knowledge and data-driven machine learning methods to overcome the disadvantages of incomplete physics knowledge and low generalization capability of data-driven models.

\subsection{Physics-Informed Machine Learning}

Physics-informed machine learning, which is also unknown as the physics-guided machine learning, is a  rising paradigm by coupling the physics scientific knowledge and data-driven machine learning methods. 
Physics-informed machine learning methods can be divided into four categories based on their different ways to couple the physics scientific knowledge and data-driven machine learning methods, including {\em physics-guided loss function, physics-guided initialization, physics-guided design of architecture, residual modeling}, and {\em hybrid physics-ML models.}
Physics-informed machine learning methods based on {\em physics-guided loss function} construct a new item in the loss function based on the physical law~\cite{willard2020integrating,karniadakis2021physics,karpatne2017theory}, which helps machine learning models to find a solution consistent with physical laws. This way is especially helpful in the scenario with sparse data, since constraining the solution consistent with physical laws helps to avoid incorrectly fitting unique mobility patterns or noise of the training dataset.
Another way to realize physics-informed machine learning is {\em physics-guided initialization}~\cite{willard2020integrating}. For example, pre-train the machine learning models with synthetic data generated by the physical knowledge~\cite{jia2019physics,jia2021physics} or simulators built on a video game physics engine~\cite{shah2018airsim}.
{\em Physics-guided design of architecture} methods propose to utilize physics scientific knowledge to construct new architectures for machine learning models, e.g., insert variables constrained by physics knowledge as the intermediate variables~\cite{Muralidhar2020PhyNet}.
Meanwhile, {\em residual modeling} methods propose to utilize machine learning models to predict the residuals made by physics-based models~\cite{Kochkove2021Machine, Yi2018Data}, and {\em hybrid physics-ML} methods operate physics-based models and machine learning models simultaneously, e.g., replace components of a physics-based model with machine learning models~\cite{parish2016paradigm,zhang2018real}.
In this paper, we also follow the paradigm of physics-informed machine learning. Specifically, we utilize the functional form of the gravity law to construct a nonlinear flow predictor, which belongs to {\em physics-guided design of architecture}.

\section{Conclusion} \label{sec:Conclusion}
Existing works on OD generation either analogize human mobility to simple physics laws or use purely data-driven ML models to model the relationship between urban structure and flow. Whereas simple physics laws cannot model complex population mobility, ML models suffer from the notorious poor generalization. In this paper, we investigate a novel physics-guided ML model called \textbf{ODGN} with a conditional GAN training strategy to solve the OD generation problem, which is crucial in many urban applications. In our approach, the strengths of physics laws and data-driven ML methods are integrated to complement each other and avoid their weaknesses. We design a discriminator in the training strategy, which determines whether the sequences come from generated OD networks or real OD networks based on the sequences sampled via probability-based random walk so that the topological features of OD networks could be taken into consideration. We evaluated the effectiveness of our method to generate OD on real data and the results have a large improvement compared to baselines.



\bibliographystyle{ACM-Reference-Format}
\bibliography{sample-base}


\begin{thebibliography}{50}


\ifx \showCODEN    \undefined \def \showCODEN     #1{\unskip}     \fi
\ifx \showDOI      \undefined \def \showDOI       #1{#1}\fi
\ifx \showISBNx    \undefined \def \showISBNx     #1{\unskip}     \fi
\ifx \showISBNxiii \undefined \def \showISBNxiii  #1{\unskip}     \fi
\ifx \showISSN     \undefined \def \showISSN      #1{\unskip}     \fi
\ifx \showLCCN     \undefined \def \showLCCN      #1{\unskip}     \fi
\ifx \shownote     \undefined \def \shownote      #1{#1}          \fi
\ifx \showarticletitle \undefined \def \showarticletitle #1{#1}   \fi
\ifx \showURL      \undefined \def \showURL       {\relax}        \fi
\providecommand\bibfield[2]{#2}
\providecommand\bibinfo[2]{#2}
\providecommand\natexlab[1]{#1}
\providecommand\showeprint[2][]{arXiv:#2}

\bibitem[Arjovsky et~al\mbox{.}(2017)]%
        {arjovsky2017wasserstein}
\bibfield{author}{\bibinfo{person}{Martin Arjovsky}, \bibinfo{person}{Soumith Chintala}, {and} \bibinfo{person}{L{\'e}on Bottou}.} \bibinfo{year}{2017}\natexlab{}.
\newblock \showarticletitle{Wasserstein generative adversarial networks}. In \bibinfo{booktitle}{\emph{International conference on machine learning}}. PMLR, \bibinfo{pages}{214--223}.
\newblock


\bibitem[Bai et~al\mbox{.}(2018)]%
        {bai2018empirical}
\bibfield{author}{\bibinfo{person}{Shaojie Bai}, \bibinfo{person}{J~Zico Kolter}, {and} \bibinfo{person}{Vladlen Koltun}.} \bibinfo{year}{2018}\natexlab{}.
\newblock \showarticletitle{An empirical evaluation of generic convolutional and recurrent networks for sequence modeling}.
\newblock \bibinfo{journal}{\emph{arXiv preprint arXiv:1803.01271}} (\bibinfo{year}{2018}).
\newblock


\bibitem[Barbosa et~al\mbox{.}(2018)]%
        {barbosa2018human}
\bibfield{author}{\bibinfo{person}{Hugo Barbosa}, \bibinfo{person}{Marc Barthelemy}, \bibinfo{person}{Gourab Ghoshal}, \bibinfo{person}{Charlotte~R James}, \bibinfo{person}{Maxime Lenormand}, \bibinfo{person}{Thomas Louail}, \bibinfo{person}{Ronaldo Menezes}, \bibinfo{person}{Jos{\'e}~J Ramasco}, \bibinfo{person}{Filippo Simini}, {and} \bibinfo{person}{Marcello Tomasini}.} \bibinfo{year}{2018}\natexlab{}.
\newblock \showarticletitle{Human mobility: Models and applications}.
\newblock \bibinfo{journal}{\emph{Physics Reports}}  \bibinfo{volume}{734} (\bibinfo{year}{2018}), \bibinfo{pages}{1--74}.
\newblock


\bibitem[Bode et~al\mbox{.}(2019)]%
        {bode2019using}
\bibfield{author}{\bibinfo{person}{Mathis Bode}, \bibinfo{person}{Michael Gauding}, \bibinfo{person}{Zeyu Lian}, \bibinfo{person}{Dominik Denker}, \bibinfo{person}{Marco Davidovic}, \bibinfo{person}{Konstantin Kleinheinz}, \bibinfo{person}{Jenia Jitsev}, {and} \bibinfo{person}{Heinz Pitsch}.} \bibinfo{year}{2019}\natexlab{}.
\newblock \showarticletitle{Using physics-informed super-resolution generative adversarial networks for subgrid modeling in turbulent reactive flows}.
\newblock \bibinfo{journal}{\emph{arXiv preprint arXiv:1911.11380}} (\bibinfo{year}{2019}).
\newblock


\bibitem[Bojchevski et~al\mbox{.}(2018)]%
        {bojchevski2018netgan}
\bibfield{author}{\bibinfo{person}{Aleksandar Bojchevski}, \bibinfo{person}{Oleksandr Shchur}, \bibinfo{person}{Daniel Z{\"u}gner}, {and} \bibinfo{person}{Stephan G{\"u}nnemann}.} \bibinfo{year}{2018}\natexlab{}.
\newblock \showarticletitle{Netgan: Generating graphs via random walks}. In \bibinfo{booktitle}{\emph{International Conference on Machine Learning}}. PMLR, \bibinfo{pages}{610--619}.
\newblock


\bibitem[Bureau(2020)]%
        {census2020}
\bibfield{author}{\bibinfo{person}{U.S.~Census Bureau}.} \bibinfo{year}{2020}\natexlab{}.
\newblock \bibinfo{title}{2020 American Community Survey}.
\newblock \bibinfo{howpublished}{U.S. Department of Commerce}.
\newblock


\bibitem[Cang et~al\mbox{.}(2018)]%
        {cang2018improving}
\bibfield{author}{\bibinfo{person}{Ruijin Cang}, \bibinfo{person}{Hechao Li}, \bibinfo{person}{Hope Yao}, \bibinfo{person}{Yang Jiao}, {and} \bibinfo{person}{Yi Ren}.} \bibinfo{year}{2018}\natexlab{}.
\newblock \showarticletitle{Improving direct physical properties prediction of heterogeneous materials from imaging data via convolutional neural network and a morphology-aware generative model}.
\newblock \bibinfo{journal}{\emph{Computational Materials Science}}  \bibinfo{volume}{150} (\bibinfo{year}{2018}), \bibinfo{pages}{212--221}.
\newblock


\bibitem[Castiglione et~al\mbox{.}(2015)]%
        {castiglione2015activity}
\bibfield{author}{\bibinfo{person}{Joe Castiglione}, \bibinfo{person}{Mark Bradley}, {and} \bibinfo{person}{John Gliebe}.} \bibinfo{year}{2015}\natexlab{}.
\newblock \bibinfo{booktitle}{\emph{Activity-based travel demand models: a primer}}.
\newblock Number SHRP 2 Report S2-C46-RR-1.
\newblock


\bibitem[Deng et~al\mbox{.}(2016)]%
        {deng2016latent}
\bibfield{author}{\bibinfo{person}{Dingxiong Deng}, \bibinfo{person}{Cyrus Shahabi}, \bibinfo{person}{Ugur Demiryurek}, \bibinfo{person}{Linhong Zhu}, \bibinfo{person}{Rose Yu}, {and} \bibinfo{person}{Yan Liu}.} \bibinfo{year}{2016}\natexlab{}.
\newblock \showarticletitle{Latent space model for road networks to predict time-varying traffic}. In \bibinfo{booktitle}{\emph{Proceedings of the 22nd ACM SIGKDD international conference on Knowledge discovery and data mining}}. \bibinfo{pages}{1525--1534}.
\newblock


\bibitem[Faghmous and Kumar(2014)]%
        {faghmous2014big}
\bibfield{author}{\bibinfo{person}{James~H Faghmous} {and} \bibinfo{person}{Vipin Kumar}.} \bibinfo{year}{2014}\natexlab{}.
\newblock \showarticletitle{A big data guide to understanding climate change: The case for theory-guided data science}.
\newblock \bibinfo{journal}{\emph{Big data}} \bibinfo{volume}{2}, \bibinfo{number}{3} (\bibinfo{year}{2014}), \bibinfo{pages}{155--163}.
\newblock


\bibitem[Jang et~al\mbox{.}(2016)]%
        {jang2016categorical}
\bibfield{author}{\bibinfo{person}{Eric Jang}, \bibinfo{person}{Shixiang Gu}, {and} \bibinfo{person}{Ben Poole}.} \bibinfo{year}{2016}\natexlab{}.
\newblock \showarticletitle{Categorical reparameterization with gumbel-softmax}.
\newblock \bibinfo{journal}{\emph{arXiv preprint arXiv:1611.01144}} (\bibinfo{year}{2016}).
\newblock


\bibitem[Jia et~al\mbox{.}(2019)]%
        {jia2019physics}
\bibfield{author}{\bibinfo{person}{Xiaowei Jia}, \bibinfo{person}{Jared Willard}, \bibinfo{person}{Anuj Karpatne}, \bibinfo{person}{Jordan Read}, \bibinfo{person}{Jacob Zwart}, \bibinfo{person}{Michael Steinbach}, {and} \bibinfo{person}{Vipin Kumar}.} \bibinfo{year}{2019}\natexlab{}.
\newblock \showarticletitle{Physics guided RNNs for modeling dynamical systems: A case study in simulating lake temperature profiles}. In \bibinfo{booktitle}{\emph{Proceedings of the 2019 SIAM International Conference on Data Mining}}. \bibinfo{pages}{558--566}.
\newblock


\bibitem[Jia et~al\mbox{.}(2021)]%
        {jia2021physics}
\bibfield{author}{\bibinfo{person}{Xiaowei Jia}, \bibinfo{person}{Jared Willard}, \bibinfo{person}{Anuj Karpatne}, \bibinfo{person}{Jordan~S Read}, \bibinfo{person}{Jacob~A Zwart}, \bibinfo{person}{Michael Steinbach}, {and} \bibinfo{person}{Vipin Kumar}.} \bibinfo{year}{2021}\natexlab{}.
\newblock \showarticletitle{Physics-guided machine learning for scientific discovery: An application in simulating lake temperature profiles}.
\newblock \bibinfo{journal}{\emph{ACM/IMS Transactions on Data Science}} \bibinfo{volume}{2}, \bibinfo{number}{3} (\bibinfo{year}{2021}), \bibinfo{pages}{1--26}.
\newblock


\bibitem[Karniadakis et~al\mbox{.}(2021)]%
        {karniadakis2021physics}
\bibfield{author}{\bibinfo{person}{George~Em Karniadakis}, \bibinfo{person}{Ioannis~G Kevrekidis}, \bibinfo{person}{Lu Lu}, \bibinfo{person}{Paris Perdikaris}, \bibinfo{person}{Sifan Wang}, {and} \bibinfo{person}{Liu Yang}.} \bibinfo{year}{2021}\natexlab{}.
\newblock \showarticletitle{Physics-informed machine learning}.
\newblock \bibinfo{journal}{\emph{Nature Reviews Physics}} \bibinfo{volume}{3}, \bibinfo{number}{6} (\bibinfo{year}{2021}), \bibinfo{pages}{422--440}.
\newblock


\bibitem[Karpatne et~al\mbox{.}(2017)]%
        {karpatne2017theory}
\bibfield{author}{\bibinfo{person}{Anuj Karpatne}, \bibinfo{person}{Gowtham Atluri}, \bibinfo{person}{James~H Faghmous}, \bibinfo{person}{Michael Steinbach}, \bibinfo{person}{Arindam Banerjee}, \bibinfo{person}{Auroop Ganguly}, \bibinfo{person}{Shashi Shekhar}, \bibinfo{person}{Nagiza Samatova}, {and} \bibinfo{person}{Vipin Kumar}.} \bibinfo{year}{2017}\natexlab{}.
\newblock \showarticletitle{Theory-guided data science: A new paradigm for scientific discovery from data}.
\newblock \bibinfo{journal}{\emph{IEEE Transactions on knowledge and data engineering}} \bibinfo{volume}{29}, \bibinfo{number}{10} (\bibinfo{year}{2017}), \bibinfo{pages}{2318--2331}.
\newblock


\bibitem[Kittel and McEuen(2018)]%
        {kittel2018kittel}
\bibfield{author}{\bibinfo{person}{Charles Kittel} {and} \bibinfo{person}{Paul McEuen}.} \bibinfo{year}{2018}\natexlab{}.
\newblock \bibinfo{booktitle}{\emph{Kittel's Introduction to Solid State Physics}}.
\newblock \bibinfo{publisher}{John Wiley \& Sons}.
\newblock


\bibitem[Kochkov et~al\mbox{.}(2021)]%
        {Kochkove2021Machine}
\bibfield{author}{\bibinfo{person}{Dmitrii Kochkov}, \bibinfo{person}{Jamie~A. Smith}, \bibinfo{person}{Ayya Alieva}, \bibinfo{person}{Qing Wang}, \bibinfo{person}{Michael~P. Brenner}, {and} \bibinfo{person}{Stephan Hoyer}.} \bibinfo{year}{2021}\natexlab{}.
\newblock \showarticletitle{Machine learning{\textendash}accelerated computational fluid dynamics}.
\newblock \bibinfo{journal}{\emph{Proceedings of the National Academy of Sciences}} \bibinfo{volume}{118}, \bibinfo{number}{21} (\bibinfo{year}{2021}).
\newblock


\bibitem[Krasnopolsky and Fox-Rabinovitz(2006)]%
        {krasnopolsky2006complex}
\bibfield{author}{\bibinfo{person}{Vladimir~M Krasnopolsky} {and} \bibinfo{person}{Michael~S Fox-Rabinovitz}.} \bibinfo{year}{2006}\natexlab{}.
\newblock \showarticletitle{Complex hybrid models combining deterministic and machine learning components for numerical climate modeling and weather prediction}.
\newblock \bibinfo{journal}{\emph{Neural Networks}} \bibinfo{volume}{19}, \bibinfo{number}{2} (\bibinfo{year}{2006}), \bibinfo{pages}{122--134}.
\newblock


\bibitem[Lenormand et~al\mbox{.}(2016)]%
        {lenormand2016systematic}
\bibfield{author}{\bibinfo{person}{Maxime Lenormand}, \bibinfo{person}{Aleix Bassolas}, {and} \bibinfo{person}{Jos{\'e}~J Ramasco}.} \bibinfo{year}{2016}\natexlab{}.
\newblock \showarticletitle{Systematic comparison of trip distribution laws and models}.
\newblock \bibinfo{journal}{\emph{Journal of Transport Geography}}  \bibinfo{volume}{51} (\bibinfo{year}{2016}), \bibinfo{pages}{158--169}.
\newblock


\bibitem[Liu et~al\mbox{.}(2020)]%
        {liu2020learning}
\bibfield{author}{\bibinfo{person}{Zhicheng Liu}, \bibinfo{person}{Fabio Miranda}, \bibinfo{person}{Weiting Xiong}, \bibinfo{person}{Junyan Yang}, \bibinfo{person}{Qiao Wang}, {and} \bibinfo{person}{Claudio Silva}.} \bibinfo{year}{2020}\natexlab{}.
\newblock \showarticletitle{Learning geo-contextual embeddings for commuting flow prediction}. In \bibinfo{booktitle}{\emph{Proceedings of the AAAI Conference on Artificial Intelligence}}, Vol.~\bibinfo{volume}{34}. \bibinfo{pages}{808--816}.
\newblock


\bibitem[Mirza and Osindero(2014)]%
        {mirza2014conditional}
\bibfield{author}{\bibinfo{person}{Mehdi Mirza} {and} \bibinfo{person}{Simon Osindero}.} \bibinfo{year}{2014}\natexlab{}.
\newblock \showarticletitle{Conditional generative adversarial nets}.
\newblock \bibinfo{journal}{\emph{arXiv preprint arXiv:1411.1784}} (\bibinfo{year}{2014}).
\newblock


\bibitem[Mohan and Gaitonde(2018)]%
        {mohan2018deep}
\bibfield{author}{\bibinfo{person}{Arvind~T Mohan} {and} \bibinfo{person}{Datta~V Gaitonde}.} \bibinfo{year}{2018}\natexlab{}.
\newblock \showarticletitle{A deep learning based approach to reduced order modeling for turbulent flow control using LSTM neural networks}.
\newblock \bibinfo{journal}{\emph{arXiv preprint arXiv:1804.09269}} (\bibinfo{year}{2018}).
\newblock


\bibitem[Muralidhar et~al\mbox{.}(2020)]%
        {Muralidhar2020PhyNet}
\bibfield{author}{\bibinfo{person}{N. Muralidhar}, \bibinfo{person}{J. Bu}, \bibinfo{person}{Z. Cao}, \bibinfo{person}{L. He}, {and} \bibinfo{person}{A. Karpatne}.} \bibinfo{year}{2020}\natexlab{}.
\newblock \showarticletitle{PhyNet: Physics Guided Neural Networks for Particle Drag Force Prediction in Assembly}.
\newblock \bibinfo{journal}{\emph{Proceedings of the 2020 SIAM International Conference on Data Mining}} (\bibinfo{year}{2020}).
\newblock


\bibitem[Newton(1833)]%
        {newton1833philosophiae}
\bibfield{author}{\bibinfo{person}{Isaac Newton}.} \bibinfo{year}{1833}\natexlab{}.
\newblock \bibinfo{booktitle}{\emph{Philosophiae naturalis principia mathematica}}. Vol.~\bibinfo{volume}{1}.
\newblock \bibinfo{publisher}{G. Brookman}.
\newblock


\bibitem[O'Gorman and Dwyer(2018)]%
        {o2018using}
\bibfield{author}{\bibinfo{person}{Paul~A O'Gorman} {and} \bibinfo{person}{John~G Dwyer}.} \bibinfo{year}{2018}\natexlab{}.
\newblock \showarticletitle{Using machine learning to parameterize moist convection: Potential for modeling of climate, climate change, and extreme events}.
\newblock \bibinfo{journal}{\emph{Journal of Advances in Modeling Earth Systems}} \bibinfo{volume}{10}, \bibinfo{number}{10} (\bibinfo{year}{2018}), \bibinfo{pages}{2548--2563}.
\newblock


\bibitem[{OpenStreetMap contributors}(2020)]%
        {OpenStreetMap}
\bibfield{author}{\bibinfo{person}{{OpenStreetMap contributors}}.} \bibinfo{year}{2020}\natexlab{}.
\newblock \bibinfo{title}{{Planet dump retrieved from https://planet.osm.org }}.
\newblock \bibinfo{howpublished}{\url{ https://www.openstreetmap.org }}.
\newblock


\bibitem[Parish and Duraisamy(2016)]%
        {parish2016paradigm}
\bibfield{author}{\bibinfo{person}{Eric~J Parish} {and} \bibinfo{person}{Karthik Duraisamy}.} \bibinfo{year}{2016}\natexlab{}.
\newblock \showarticletitle{A paradigm for data-driven predictive modeling using field inversion and machine learning}.
\newblock \bibinfo{journal}{\emph{J. Comput. Phys.}}  \bibinfo{volume}{305} (\bibinfo{year}{2016}), \bibinfo{pages}{758--774}.
\newblock


\bibitem[Pourebrahim et~al\mbox{.}(2019)]%
        {pourebrahim2019trip}
\bibfield{author}{\bibinfo{person}{Nastaran Pourebrahim}, \bibinfo{person}{Selima Sultana}, \bibinfo{person}{Amirreza Niakanlahiji}, {and} \bibinfo{person}{Jean-Claude Thill}.} \bibinfo{year}{2019}\natexlab{}.
\newblock \showarticletitle{Trip distribution modeling with Twitter data}.
\newblock \bibinfo{journal}{\emph{Computers, Environment and Urban Systems}}  \bibinfo{volume}{77} (\bibinfo{year}{2019}), \bibinfo{pages}{101354}.
\newblock


\bibitem[Raccuglia et~al\mbox{.}(2016)]%
        {raccuglia2016machine}
\bibfield{author}{\bibinfo{person}{Paul Raccuglia}, \bibinfo{person}{Katherine~C Elbert}, \bibinfo{person}{Philip~DF Adler}, \bibinfo{person}{Casey Falk}, \bibinfo{person}{Malia~B Wenny}, \bibinfo{person}{Aurelio Mollo}, \bibinfo{person}{Matthias Zeller}, \bibinfo{person}{Sorelle~A Friedler}, \bibinfo{person}{Joshua Schrier}, {and} \bibinfo{person}{Alexander~J Norquist}.} \bibinfo{year}{2016}\natexlab{}.
\newblock \showarticletitle{Machine-learning-assisted materials discovery using failed experiments}.
\newblock \bibinfo{journal}{\emph{Nature}} \bibinfo{volume}{533}, \bibinfo{number}{7601} (\bibinfo{year}{2016}), \bibinfo{pages}{73--76}.
\newblock


\bibitem[Reichstein et~al\mbox{.}(2019)]%
        {reichstein2019deep}
\bibfield{author}{\bibinfo{person}{Markus Reichstein}, \bibinfo{person}{Gustau Camps-Valls}, \bibinfo{person}{Bjorn Stevens}, \bibinfo{person}{Martin Jung}, \bibinfo{person}{Joachim Denzler}, \bibinfo{person}{Nuno Carvalhais}, {et~al\mbox{.}}} \bibinfo{year}{2019}\natexlab{}.
\newblock \showarticletitle{Deep learning and process understanding for data-driven Earth system science}.
\newblock \bibinfo{journal}{\emph{Nature}} \bibinfo{volume}{566}, \bibinfo{number}{7743} (\bibinfo{year}{2019}), \bibinfo{pages}{195--204}.
\newblock


\bibitem[Robinson and Dilkina(2018)]%
        {robinson2018machine}
\bibfield{author}{\bibinfo{person}{Caleb Robinson} {and} \bibinfo{person}{Bistra Dilkina}.} \bibinfo{year}{2018}\natexlab{}.
\newblock \showarticletitle{A machine learning approach to modeling human migration}. In \bibinfo{booktitle}{\emph{Proceedings of the 1st ACM SIGCAS Conference on Computing and Sustainable Societies}}. \bibinfo{pages}{1--8}.
\newblock


\bibitem[Ruiter(1967)]%
        {ruiter1967toward}
\bibfield{author}{\bibinfo{person}{Earl~R Ruiter}.} \bibinfo{year}{1967}\natexlab{}.
\newblock \showarticletitle{Toward a better understanding of the intervening opportunities model}.
\newblock \bibinfo{journal}{\emph{Transportation Research}} \bibinfo{volume}{1}, \bibinfo{number}{1} (\bibinfo{year}{1967}), \bibinfo{pages}{47--56}.
\newblock


\bibitem[Saberi et~al\mbox{.}(2017)]%
        {saberi2017complex}
\bibfield{author}{\bibinfo{person}{Meead Saberi}, \bibinfo{person}{Hani~S Mahmassani}, \bibinfo{person}{Dirk Brockmann}, {and} \bibinfo{person}{Amir Hosseini}.} \bibinfo{year}{2017}\natexlab{}.
\newblock \showarticletitle{A complex network perspective for characterizing urban travel demand patterns: graph theoretical analysis of large-scale origin--destination demand networks}.
\newblock \bibinfo{journal}{\emph{Transportation}} \bibinfo{volume}{44}, \bibinfo{number}{6} (\bibinfo{year}{2017}), \bibinfo{pages}{1383--1402}.
\newblock


\bibitem[Salha et~al\mbox{.}(2019)]%
        {salha2019gravity}
\bibfield{author}{\bibinfo{person}{Guillaume Salha}, \bibinfo{person}{Stratis Limnios}, \bibinfo{person}{Romain Hennequin}, \bibinfo{person}{Viet-Anh Tran}, {and} \bibinfo{person}{Michalis Vazirgiannis}.} \bibinfo{year}{2019}\natexlab{}.
\newblock \showarticletitle{Gravity-Inspired Graph Autoencoders for Directed Link Prediction. CoRR abs/1905.09570 (2019)}.
\newblock \bibinfo{journal}{\emph{arXiv preprint arXiv:1905.09570}} (\bibinfo{year}{2019}).
\newblock


\bibitem[Schleder et~al\mbox{.}(2019)]%
        {schleder2019dft}
\bibfield{author}{\bibinfo{person}{Gabriel~R Schleder}, \bibinfo{person}{Antonio~CM Padilha}, \bibinfo{person}{Carlos~Mera Acosta}, \bibinfo{person}{Marcio Costa}, {and} \bibinfo{person}{Adalberto Fazzio}.} \bibinfo{year}{2019}\natexlab{}.
\newblock \showarticletitle{From DFT to machine learning: recent approaches to materials science--a review}.
\newblock \bibinfo{journal}{\emph{Journal of Physics: Materials}} \bibinfo{volume}{2}, \bibinfo{number}{3} (\bibinfo{year}{2019}), \bibinfo{pages}{032001}.
\newblock


\bibitem[Shah et~al\mbox{.}(2018)]%
        {shah2018airsim}
\bibfield{author}{\bibinfo{person}{Shital Shah}, \bibinfo{person}{Debadeepta Dey}, \bibinfo{person}{Chris Lovett}, {and} \bibinfo{person}{Ashish Kapoor}.} \bibinfo{year}{2018}\natexlab{}.
\newblock \showarticletitle{Airsim: High-fidelity visual and physical simulation for autonomous vehicles}. In \bibinfo{booktitle}{\emph{Field and service robotics}}. \bibinfo{pages}{621--635}.
\newblock


\bibitem[Shi et~al\mbox{.}(2020)]%
        {shi2020predicting}
\bibfield{author}{\bibinfo{person}{Hongzhi Shi}, \bibinfo{person}{Quanming Yao}, \bibinfo{person}{Qi Guo}, \bibinfo{person}{Yaguang Li}, \bibinfo{person}{Lingyu Zhang}, \bibinfo{person}{Jieping Ye}, \bibinfo{person}{Yong Li}, {and} \bibinfo{person}{Yan Liu}.} \bibinfo{year}{2020}\natexlab{}.
\newblock \showarticletitle{Predicting Origin-Destination Flow via Multi-Perspective Graph Convolutional Network}. In \bibinfo{booktitle}{\emph{2020 IEEE 36th International Conference on Data Engineering (ICDE)}}. IEEE, \bibinfo{pages}{1818--1821}.
\newblock


\bibitem[Simini et~al\mbox{.}(2021)]%
        {simini2021deep}
\bibfield{author}{\bibinfo{person}{Filippo Simini}, \bibinfo{person}{Gianni Barlacchi}, \bibinfo{person}{Massimilano Luca}, {and} \bibinfo{person}{Luca Pappalardo}.} \bibinfo{year}{2021}\natexlab{}.
\newblock \showarticletitle{A Deep Gravity model for mobility flows generation}.
\newblock \bibinfo{journal}{\emph{Nature communications}} \bibinfo{volume}{12}, \bibinfo{number}{1} (\bibinfo{year}{2021}), \bibinfo{pages}{1--13}.
\newblock


\bibitem[Simini et~al\mbox{.}(2012)]%
        {simini2012universal}
\bibfield{author}{\bibinfo{person}{Filippo Simini}, \bibinfo{person}{Marta~C Gonz{\'a}lez}, \bibinfo{person}{Amos Maritan}, {and} \bibinfo{person}{Albert-L{\'a}szl{\'o} Barab{\'a}si}.} \bibinfo{year}{2012}\natexlab{}.
\newblock \showarticletitle{A universal model for mobility and migration patterns}.
\newblock \bibinfo{journal}{\emph{Nature}} \bibinfo{volume}{484}, \bibinfo{number}{7392} (\bibinfo{year}{2012}), \bibinfo{pages}{96--100}.
\newblock


\bibitem[Sobral et~al\mbox{.}(2021)]%
        {Sobral2021Knowledge}
\bibfield{author}{\bibinfo{person}{T. Sobral}, \bibinfo{person}{T. Galvao}, {and} \bibinfo{person}{J. Borges}.} \bibinfo{year}{2021}\natexlab{}.
\newblock \showarticletitle{Knowledge-Assisted Visualization of Multi-Level Origin-Destination Flows Using Ontologies}.
\newblock \bibinfo{journal}{\emph{IEEE Transactions on Intelligent Transportation Systems}} \bibinfo{volume}{PP}, \bibinfo{number}{99} (\bibinfo{year}{2021}), \bibinfo{pages}{1--10}.
\newblock


\bibitem[Spadon et~al\mbox{.}(2019)]%
        {spadon2019reconstructing}
\bibfield{author}{\bibinfo{person}{Gabriel Spadon}, \bibinfo{person}{Andre~CPLF de Carvalho}, \bibinfo{person}{Jose~F Rodrigues-Jr}, {and} \bibinfo{person}{Luiz~GA Alves}.} \bibinfo{year}{2019}\natexlab{}.
\newblock \showarticletitle{Reconstructing commuters network using machine learning and urban indicators}.
\newblock \bibinfo{journal}{\emph{Scientific reports}} \bibinfo{volume}{9}, \bibinfo{number}{1} (\bibinfo{year}{2019}), \bibinfo{pages}{1--13}.
\newblock


\bibitem[Veli{\v{c}}kovi{\'c} et~al\mbox{.}(2017)]%
        {velivckovic2017graph}
\bibfield{author}{\bibinfo{person}{Petar Veli{\v{c}}kovi{\'c}}, \bibinfo{person}{Guillem Cucurull}, \bibinfo{person}{Arantxa Casanova}, \bibinfo{person}{Adriana Romero}, \bibinfo{person}{Pietro Lio}, {and} \bibinfo{person}{Yoshua Bengio}.} \bibinfo{year}{2017}\natexlab{}.
\newblock \showarticletitle{Graph attention networks}.
\newblock \bibinfo{journal}{\emph{arXiv preprint arXiv:1710.10903}} (\bibinfo{year}{2017}).
\newblock


\bibitem[Wang et~al\mbox{.}(2019)]%
        {wang2019origin}
\bibfield{author}{\bibinfo{person}{Yuandong Wang}, \bibinfo{person}{Hongzhi Yin}, \bibinfo{person}{Hongxu Chen}, \bibinfo{person}{Tianyu Wo}, \bibinfo{person}{Jie Xu}, {and} \bibinfo{person}{Kai Zheng}.} \bibinfo{year}{2019}\natexlab{}.
\newblock \showarticletitle{Origin-destination matrix prediction via graph convolution: a new perspective of passenger demand modeling}. In \bibinfo{booktitle}{\emph{Proceedings of the 25th ACM SIGKDD International Conference on Knowledge Discovery \& Data Mining}}. \bibinfo{pages}{1227--1235}.
\newblock


\bibitem[Willard et~al\mbox{.}(2020)]%
        {willard2020integrating}
\bibfield{author}{\bibinfo{person}{Jared Willard}, \bibinfo{person}{Xiaowei Jia}, \bibinfo{person}{Shaoming Xu}, \bibinfo{person}{Michael Steinbach}, {and} \bibinfo{person}{Vipin Kumar}.} \bibinfo{year}{2020}\natexlab{}.
\newblock \showarticletitle{Integrating physics-based modeling with machine learning: A survey}.
\newblock \bibinfo{journal}{\emph{arXiv preprint arXiv:2003.04919}} \bibinfo{volume}{1}, \bibinfo{number}{1} (\bibinfo{year}{2020}), \bibinfo{pages}{1--34}.
\newblock


\bibitem[Xiao et~al\mbox{.}(2019)]%
        {xiao2019reduced}
\bibfield{author}{\bibinfo{person}{Dunhui Xiao}, \bibinfo{person}{CE Heaney}, \bibinfo{person}{L Mottet}, \bibinfo{person}{F Fang}, \bibinfo{person}{W Lin}, \bibinfo{person}{IM Navon}, \bibinfo{person}{Y Guo}, \bibinfo{person}{OK Matar}, \bibinfo{person}{AG Robins}, {and} \bibinfo{person}{CC Pain}.} \bibinfo{year}{2019}\natexlab{}.
\newblock \showarticletitle{A reduced order model for turbulent flows in the urban environment using machine learning}.
\newblock \bibinfo{journal}{\emph{Building and Environment}}  \bibinfo{volume}{148} (\bibinfo{year}{2019}), \bibinfo{pages}{323--337}.
\newblock


\bibitem[Yao et~al\mbox{.}(2020)]%
        {Yao2020Spatial}
\bibfield{author}{\bibinfo{person}{X. Yao}, \bibinfo{person}{Y. Gao}, \bibinfo{person}{D. Zhu}, \bibinfo{person}{E. Manley}, {and} \bibinfo{person}{Y. Liu}.} \bibinfo{year}{2020}\natexlab{}.
\newblock \showarticletitle{Spatial Origin-Destination Flow Imputation Using Graph Convolutional Networks}.
\newblock \bibinfo{journal}{\emph{IEEE Transactions on Intelligent Transportation Systems}} \bibinfo{volume}{PP}, \bibinfo{number}{99} (\bibinfo{year}{2020}), \bibinfo{pages}{1--11}.
\newblock


\bibitem[Yazdani et~al\mbox{.}(2020)]%
        {yazdani2020systems}
\bibfield{author}{\bibinfo{person}{Alireza Yazdani}, \bibinfo{person}{Lu Lu}, \bibinfo{person}{Maziar Raissi}, {and} \bibinfo{person}{George~Em Karniadakis}.} \bibinfo{year}{2020}\natexlab{}.
\newblock \showarticletitle{Systems biology informed deep learning for inferring parameters and hidden dynamics}.
\newblock \bibinfo{journal}{\emph{PLoS computational biology}} \bibinfo{volume}{16}, \bibinfo{number}{11} (\bibinfo{year}{2020}), \bibinfo{pages}{e1007575}.
\newblock


\bibitem[Yi et~al\mbox{.}(2018)]%
        {Yi2018Data}
\bibfield{author}{\bibinfo{person}{W.~Z. Yi}, \bibinfo{person}{V. Pantelis}, \bibinfo{person}{K. Petros}, \bibinfo{person}{S. Themistoklis}, {and} \bibinfo{person}{D. Daniel}.} \bibinfo{year}{2018}\natexlab{}.
\newblock \showarticletitle{Data-assisted reduced-order modeling of extreme events in complex dynamical systems}.
\newblock \bibinfo{journal}{\emph{Plos One}} \bibinfo{volume}{13}, \bibinfo{number}{5} (\bibinfo{year}{2018}).
\newblock


\bibitem[Zhang et~al\mbox{.}(2018)]%
        {zhang2018real}
\bibfield{author}{\bibinfo{person}{Liang Zhang}, \bibinfo{person}{Gang Wang}, {and} \bibinfo{person}{Georgios~B Giannakis}.} \bibinfo{year}{2018}\natexlab{}.
\newblock \showarticletitle{Real-time power system state estimation via deep unrolled neural networks}. In \bibinfo{booktitle}{\emph{2018 IEEE Global Conference on Signal and Information Processing (GlobalSIP)}}. \bibinfo{pages}{907--911}.
\newblock


\bibitem[Zipf(1946)]%
        {zipf1946p}
\bibfield{author}{\bibinfo{person}{George~Kingsley Zipf}.} \bibinfo{year}{1946}\natexlab{}.
\newblock \showarticletitle{The P 1 P 2/D hypothesis: on the intercity movement of persons}.
\newblock \bibinfo{journal}{\emph{American sociological review}} \bibinfo{volume}{11}, \bibinfo{number}{6} (\bibinfo{year}{1946}), \bibinfo{pages}{677--686}.
\newblock


\end{thebibliography}



\end{document}